\documentclass{article}
\usepackage{spconf,amsmath,graphicx}
\usepackage{multirow}
\usepackage{graphicx}
\usepackage{caption}
\usepackage{subcaption}
\usepackage{booktabs}   
\usepackage{amssymb}

\title{A General Multi-Task Learning Framework to Leverage Text Data for Speech to Text Tasks}
\name{Yun Tang$^1$, Juan Pino$^1$, Changhan Wang$^1$, Xutai Ma$^{1,2}$, Dmitriy Genzel$^1$}
\address{
 	$^1$Facebook AI, USA, $\;\;\;\;\;\;\;\;$ $^2$Johns Hopkins University, USA \\
 	\{yuntang,juancarabina,changhan,dgenzel\}@fb.com, $\;\;\;\;$ xutai\_\thinspace ma@jhu.edu
}

\begin{document}
%
\maketitle
\begin{abstract}
Attention-based sequence-to-sequence modeling provides a powerful and elegant solution for applications that need to map one sequence to a different sequence. 
Its success heavily relies on the availability of large amounts of training data. 
This presents a challenge for speech applications where labelled speech data is very expensive to obtain, such as automatic speech recognition (ASR) and speech translation (ST). 
In this study, we propose a general multi-task learning framework to leverage text data for ASR and ST tasks.
Two auxiliary tasks, a denoising autoencoder task and machine translation task, are proposed to be co-trained with ASR and ST tasks respectively. 
We demonstrate that representing text input as phoneme sequences can reduce the difference between speech and text inputs, and enhance the knowledge transfer from text corpora to the speech to text tasks. 
Our experiments show that the proposed method achieves a relative 10$\sim$15\% word error rate reduction on the English \textsc{Librispeech} task compared with our baseline, and improves the speech translation quality on the \textsc{MuST-C} tasks by 3.6$\sim$9.2 BLEU.
\end{abstract}
\begin{keywords}
Multi-task learning, speech recognition, speech translation
\end{keywords}
\section{Introduction}
\label{sec:intro}
Attention-based encoder-decoder modeling is a natural and powerful paradigm for speech to text tasks, such as automatic speech recognition (ASR)
and speech translation (ST), and that has led to significant progress ~\cite{Chorowski2015AttentionBasedMF,Chan2016ListenAA,Berard2016ListenAT,Weiss2017SequencetoSequenceMC,Inaguma2020ESPnetSTAS,Wang2020fairseqSF}.  However, it relies on large amounts of supervised speech data, which is expensive to transcribe and translate. 
In addition, the amount of speech transcripts and speech translation labels is dwarfed by the amount of text data available for language model (LM) and machine translation (MT) training.
For example, the number of text tokens used for LM modeling is two orders of magnitude larger than the number of tokens from the corresponding speech corpus in the \textsc{Librispeech} data corpus~\cite{Panayotov2015LibrispeechAA}, as shown in Table~\ref{tab:data}.  

Attention-based encoder-decoder models are not designed to incorporate heterogeneous inputs and cannot benefit from large amounts of low cost text data directly in speech applications.
As expected, performance gaps can still be observed between attention based encoder-decoder systems and conventional systems with multiple components~\cite{Lscher2019RWTHAS,Niehues2019TheI2,salesky2020phone}. 
In order to alleviate the data scarcity issue, different approaches have been studied, including acoustic~\cite{Ko2017ASO,park2019specaugment} and linguistic aspects~\cite{Chorowski2016TowardsBD,rosenberg2019speech}. 
LM is the most commonly used method to integrate linguistic information into ASR. 
Prior work focuses on building LM with monolingual text data, and then integrate LM or transfer knowledge from it into the decoder~\cite{Chorowski2016TowardsBD,Sriram2017ColdFT,Bai2019LearnSF}.  
\cite{Jia2018LeveragingWS} generate synthetic data from text to augment speech training corpus.
Another direction is to leverage text data directly during training through multitask learning~\cite{Bahar2019ACS,Huang2020LeveragingUT,Renduchintala2018MultiModalDA}.
\cite{Huang2020LeveragingUT} use a common representation space to learn correspondences between different modalities for spoken language understanding.
\cite{Renduchintala2018MultiModalDA} propose multi-modal data augmentation to jointly train text and speech for ASR. 
\cite{Weiss2017SequencetoSequenceMC,Anastasopoulos2018TiedML} are focused on ST tasks and trained with an ASR system together, where ASR is used as an auxiliary task. Hence, those methods cannot be applied back to ASR systems.

In this study, we focus on leveraging text data to improve linguistic modeling ability in speech to text tasks. We propose a general framework to leverage text data for ASR and ST tasks. 
A denoising autoencoder task~\cite{Lample2018UnsupervisedMT,lewis2019bart}  is introduced to be jointly trained with the ASR task with monolingual data, while a machine translation task is co-trained with  ST task  with  parallel  data.
Text  input  is  represented  as  spoken form using phoneme sequence and it effectively reduces the difference between speech input and text input.
We also carefully study different design choices for the joint training system, including strategies to share the text and speech encoders and comparing the joint training system with models initialized from pre-trained components.  
Our experiments show the proposed joint training systems can effectively reduce word error rate (WER) for the ASR task by 10\% to 15\% and improve BLEU score by 3.6$\sim$9.2 for ST tasks. 

\begin{figure*}
    \centering
    \includegraphics[width=0.65\textwidth]{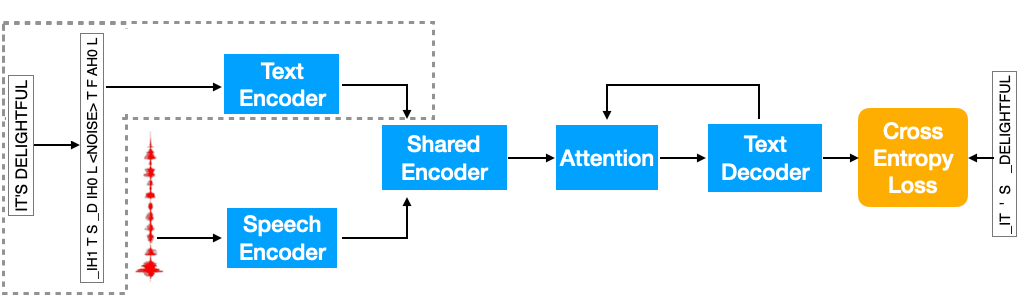}
    \caption{Joint training framework. Text encoder (within dashed box) is dropped during inference.}
    \label{fig:joint_train}
\end{figure*}
\section{Model}
The speech and text joint training framework is described in Fig. \ref{fig:joint_train}. It consists of three components: speech encoder, text encoder (within the dashed box) and decoder.  
Two types of input data are fed into the model: speech feature $X = (x_1, x_2 \cdot\cdot\cdot, x_N), x_n \in\mathcal{R}^{d_s}$ and text form tokens $Y = (y_1,y_2, \cdot\cdot\cdot, y_M)$.  The model output is a token sequence $W=(w_1,w_2,\cdot\cdot\cdot,w_K)$, where $d_s$ is the speech feature dimension, 
$N$, $M$ and $K$ are the number of speech feature frames, number of text form tokens, number of transcription or translation tokens. 
The text encoder is only used during training and dropped during inference. 
The training minimizes the loss $L$ from two sub-tasks, e.g., the ST (or ASR) task and the MT (or denoising autoencoder) task, as below : 
\begin{eqnarray}
L\! =\!-\!\texttt{log} P(W| X, \theta^s_{enc},\theta_{dec})\! -\!\texttt{log} P(W| Y, \theta^t_{enc},\theta_{dec})  
\end{eqnarray}
where $\theta^s_{enc},\theta^t_{enc}$ and $\theta_{dec}$ are parameters for speech encoder, text encoder and  decoder respectively.

\subsection{Transformer Encoders and Decoder}
Two encoders and decoder are all transformer based models~\cite{Vaswani2017AttentionIA}.   Following~\cite{Synnaeve2019EndtoendAF}, the speech encoder has an extra subsampling module, which consists of two 1-dimensional convolutional layers and each is with kernel width 3 and stride 2.  The subsampling module outputs are combined with sinusoidal positional embeddings and fed into the following transformer layers. The text encoder and decoder consist of standard transformer layers.  
We use pre-layer normalization for more stable training~\cite{xiong2020on}.

\subsection{Input Text Representation}
The input text to the joint training model comes from two sources: the transcripts from the speech training corpora, and  text data from non-speech corpora. The text input is presented as phoneme sequence instead of word tokens or subword tokens~\cite{Renduchintala2018MultiModalDA}. Phoneme sequence is the spoken form representation of the original text and is easier to be mapped to the corresponding speech input. 
The conversion from text to phoneme representation can be done through a grapheme to phoneme system or simply a dictionary lookup.
The phoneme set used in this study is based on the Carnegie Mellon Pronouncing Dictionary with 39 phonemes. Vowels carry a lexical stress marker. We further extend the phoneme set by identifying the first phoneme in the word with additional ``\_'' mark, which is similar to the notation in sentence piece process. 

\subsection{Auxiliary Text Tasks}
The MT task is chosen as an auxiliary task to be jointly trained with the ST task. The input tokens are phoneme sequences converted from the corresponding source text and the target is represented as subword units derived from the corresponding translation labels.

In the ASR task, the auxiliary task maps the input phoneme sequence to the corresponding subword tokens derived from transcripts. However, the task is too simple since 76\% of words in the Carnegie Mellon Pronouncing Dictionary can be recovered from a phoneme sequence deterministically. It brings little help to the ASR task as shown in the experiments in section~\ref{sec:exp_rst}. Hence, we modify this task to a denoising autoencoder task~\cite{Lample2018UnsupervisedMT,lewis2019bart} and part of the input phoneme tokens are masked as ``$\langle$NOISE$\rangle$". The decoder has to infer the target word based on the phonemes of this word as well as neighbouring context. Hence, the monolingual text data can be integrated into sequence to sequence ASR modeling naturally and effectively. 



Fig.~\ref{fig:joint_train} shows the ASR input/output. The input text is ``It's delightful'' and the corresponding target subword sequence encoded with SentencePiece
is ``\_IT ' S \_DELIGHTFUL''.
The phoneme sequence corresponding to input text is ``\_IH1 T S \_D IH0 L AY1 T F AH0 L''.  
Phoneme ``AY1'' is randomly selected and masked with ``$\langle$NOISE$\rangle$'' token, and the phoneme sequence fed to the model becomes ``\_IH1 T S \_D IH0 L $\langle$NOISE$\rangle$ T F AH0 L''.


\section{Experimental Settings}
\subsection{Data}
We conduct experiments on two datasets,  \textsc{Librispeech}~\cite{Panayotov2015LibrispeechAA} and \textsc{MuST-C}~\cite{Gangi2019MuSTCAM}.
The detailed training data statistics are presented in Table \ref{tab:data}. The second column is the total number of hours for the speech 
training data. The third and fifth columns are the number of (source) words.
\begin{table}
    \centering
    \small
    \begin{tabular}{l|c|c||c|c}
    \toprule
    Speech  & hours &  \#W(m) & Text & \#W(m) \\
    \hline
    LibriSpeech & 960 & 9.4 & Gutenberg & 803.3 \\ 
    \hline
    \hline
    MuST-C & & & & \\
    $ \;\; \;\; $  EN-DE & 408 & 4.2 & WMT17 & 105.2 \\
    $ \;\; \;\; $  EN-ES & 504 & 5.2 & WMT13  & 369.4 \\
    $ \;\; \;\; $  EN-FR & 492 & 5.1 & WMT14 & 1,018.5 \\
    \bottomrule
    \end{tabular}
    \caption{ Data statistics for ST and ASR training sets. ``\#W(m)'' stands for ``number of words (million)''.}\label{tab:data}
\end{table}

\noindent\textbf{ASR datasets}: 
ASR is evaluated on the \textsc{Librispeech} dataset and the co-training text data is the language model training data coming with the \textsc{Librispeech} dataset. 

\noindent\textbf{ST datasets}: ST is evaluated on three language pairs: English-German (EN-DE), English-Spanish (EN-ES) and English-French (EN-FR) on the tst-COMMON test set.   WMT parallel data is used as text training corpus. 
Case-sensitive detokenized BLEU is reported by \textsc{sacrebleu}. 

 Target subword units are learned from SentencePiece
 with vocabulary size 10k and full character coverage on all training text data. The grapheme to phoneme conversion for the input text is done through the ``g2p\_en'' Python package~\cite{Lee2018LearningPF}. The input phoneme vocabulary size is 134. 
Input speech is represented as 80D log mel-filterbank coefficients computed every 10ms with a 25ms window.  Global channel mean and variance normalization is applied to the input speech features. SpecAugment~\cite{park2019specaugment} is employed to augment audio data for model training. LD and LB policies from~\cite{park2019specaugment} are applied to \textsc{Librispeech} and \textsc{MuST-C} tasks respectively.

\begin{table}
    \centering
    \small
    \begin{tabular}{l|c| c|c|c|c}
    \toprule
    \multirow{2}{*}{Data set} &\#params  &\multicolumn{2}{c|}{Dev} & \multicolumn{2}{c}{Test}  \\
    \cline{3-6}
       & (m) & clean & other & clean & other     \\
    \hline
    LAS~\cite{park2019specaugment} & 360 & - & - &2.8 &6.8  \\ 
    Transformer~\cite{Synnaeve2019EndtoendAF} & 270 & 2.5 & 6.7 & 2.9 & 7.0 \\
    \hline
    \hline 
    Transformer (M) & 76 & 3.5 & 8.1 & 3.7 & 8.1  \\
    Joint Training (M)  & 76 & 3.0 & 7.4 & 3.3 & 7.6 \\
    \hline
    \hline
    Transformer (L) & 161 & 3.3 & 7.9 & 3.6 & 8.0  \\
    Joint Training (L) & 161 & 2.8 & 7.0 & 3.1 & 7.2 \\
    \bottomrule
    \end{tabular}
    \caption{ WER results on Librispeech }\label{tab:asr}
\end{table}
\subsection{Model Setup}
In all experiments, the speech encoder has 12 transformer layers while the text encoder and decoder have 6 transformer layers. Three model configurations are examined: small, medium and large. The small configuration has a word embedding size of 256 and transformer middle layer dimension 2048; medium configuration has word embedding size equal to 512 and the middle layer dimension 2048; the large configuration sets word embedding size to 768 and middle layer dimension to 3072.  The \textsc{MuST-C} models are using the small configuration, while the medium and large configurations are used for the \textsc{Librispeech} models. If not specifically mentioned, the medium configuration is used by default in \textsc{Librispeech} experiments. We use the  Adam  optimizer 
with a learning rate 0.001 in all experiments. Label smoothing and dropout rate are both set to 0.1. The \textsc{Librispeech} models are trained with 240 epochs using 16 GPUs and the \textsc{MuST-C} models are trained with 200 epochs using 8 GPUs. The batch size is 40,000 frames for speech samples and 20,000 tokens for text samples per GPU. Speech input and text input are used to update the model alternatively.  The models are trained with Fairseq~\cite{Wang2020fairseqSF}. 
The last 10 checkpoints are averaged for inference with beam size 5.

\begin{table}
    \centering
    \small
    \begin{tabular}{l|c|c|c|c}
    \toprule
    {Data corpus} &\#params(m)& {EN-DE} & {EN-ES} &{EN-FR} \\
    \hline
    Transformer~\cite{gangi2019onetomany} & 30& 17.7 &   20.9  &   26.5  \\
    Transformer~\cite{Inaguma2020ESPnetSTAS} & - & 22.9 & 28.0 & 32.7 \\
    Transformer~\cite{Pino2020SelfTrainingFE} & 435 & 25.2 & - & 34.5 \\
    \hline\hline
    Transformer &31 & 20.3 &  19.4     &    25.3   \\
    Joint Training & 31 & 23.9 &   28.6   &   33.1    \\
    \bottomrule
    \end{tabular}
    \caption{ BLEU results of three language pairs on the MuST-C tst-COMMON. }\label{tab:ast}
\end{table}
\section{Experimental Results}\label{sec:exp_rst}
We present our main ASR results in Table~\ref{tab:asr}. The top of the table shows results from literature. ``Transformer"  and ``Joint Training" are results from the baseline and the corresponding jointly trained system. In the jointly trained system, 20\% of input tokens are masked. Two configurations: medium (row 3-4) and large (row 5-6 ), are studied.
For both configurations, the multi-task system outperforms the baseline system at all 4 test sets. The large configuration achieves the best results with relative WER reductions varied from 10\% to 15\%.

Table~\ref{tab:ast} demonstrates the main results on three ST tasks. The models are trained with the small configuration. 
The results from baseline (row 4) are comparable to results in~\cite{gangi2019onetomany}, though all baselines are trained from 
scratch with random initialization. Row 5 demonstrates results from the jointly trained models with extra WMT parallel text. The
performance is improved significantly that 3.6 to 9.2 BLEU score increases are observed in different language pairs. 

\begin{table}
    \centering
    \small
    \begin{tabular}{l|c|c|c}
    \toprule
    \multirow{3}{*}{Data set} & \multicolumn{2}{c|}{Librispeech Dev} & {MuST-C EN-DE}  \\
     & \multicolumn{2}{c|}{(WER)} & (BLEU) \\
    \cline{2-4}
        & clean & other & tst-COMMON      \\
        \hline
        None & 3.5 & 8.1 & 20.3 \\
    \hline
    Character &3.8  & 8.6 & 23.1 \\ 
    \hline
    Phoneme & 3.0 & 7.4 & 23.9    \\
    \bottomrule
    \end{tabular}
    \caption{ Comparison of the input text representations.}\label{tab:token}
\end{table}
\noindent\textbf{Impact of Input Token Representations}
In Table~\ref{tab:token}, we compared the phoneme representation with character based representation on both ASR and ST tasks. ``None" represents results from a system without text input. 
Results from the character based text representation are listed in row ``Character". For the ASR task, the WERs are even higher than results from the baseline system trained with speech data only.  Though the ST result using character representation is better than the baseline, it is 0.8 BLEU lower than the phoneme representation based system. It is clear that organizing the input text in spoken form is critical for the speech-text multi-task learning, and it makes the knowledge transfer from text to speech more effective.

\begin{table}
    \centering
    \small
    \begin{tabular}{l|c|c|c}
    \toprule
    \multirow{3}{*}{Data set} &\multicolumn{2}{c|}{Librispeech Dev} & MuST-C EN-DE  \\
    & \multicolumn{2}{c|}{(WER)} & (BLEU) \\
    \cline{2-4}
      & clean & other & tst-COMMON \\
    \hline 
    No Share & 3.0 & 7.4  & 22.7   \\
    \hline
    Share  & 3.0 & 7.4 & 23.9 \\
    \bottomrule
    \end{tabular}
    \caption{ Comparison of encoder parameter sharing strategies}\label{tab:sharing}
\end{table}

\noindent\textbf{Encoder Parameter Sharing Strategies}
  Speech to text tasks obtain linguistic information from text corpora via a shared decoder. It is natural to ask if sharing parameters between two encoders would improve performance further.  We compare systems with shared encoder (``Share") and dedicated encoder (``No Share") in Table~\ref{tab:sharing}. Shared encoder means parameters in all 6 transformer layers in the text encoder are shared with the last 6 layers of the speech encoder.
Sharing parameters between encoders is helpful for the \textsc{MuST-C} task. 1.2 BLEU  decrease is observed without parameter sharing between two encoders. On the other hand, the ASR task is not sensitive to the parameter sharing that 
different strategies give comparable results. 

\noindent\textbf{Impact of Masking}
In Table~\ref{tab:masking}, different input text masking ratios are examined for both ASR and ST tasks. Without masking (row 1), the jointly trained ASR model performs no better than the  Transformer baseline (as shown in Table~\ref{tab:token} row ``None"). The best WER is achieved when 20\% input tokens are masked.
It confirms our assumption that mapping from phoneme sequence to the corresponding (sub)word sequence is too simple that the model can accomplish it without learning much of the language context. 
Masking some phonemes forces the model to learn better language context representation.
For the ST task, masking degrades speech translation performance.  We hypothesis that the English to German translation task is challenging, and the decoder needs to understand language context to associate the input phoneme sequence with the target subword sequence. 

\begin{table}
    \centering
    \small
    \begin{tabular}{l|c|c|c}
    \toprule
    \multirow{2}{*}{Data set} &\multicolumn{2}{c|}{Librispeech Dev} & MuST-C EN-DE  \\
    & \multicolumn{2}{c|}{(WER)} & (BLEU) \\
    \cline{2-4}
     Masking ratio & clean & other & tst-COMMON \\
    \hline 
    0.0 & 3.3 & 8.2 &  23.9  \\
    \hline
    0.1 & 2.9 & 7.6   & 23.8  \\
    \hline
    0.2 & 3.0 & 7.4   & 23.2     \\
    \hline
    0.3 & 3.0  & 7.8  & 23.1 \\
    \bottomrule
    \end{tabular}
    \caption{ Comparison of different masking ratios.}\label{tab:masking}
\end{table}

\noindent\textbf{Comparison with  Pre-Trained Models}
Compared to multi-task learning, another important direction is to learn different tasks individually, and then finetune the pre-trained model for downstream tasks. We compare these two approaches in Table~\ref{tab:pre-train}. 
Two pre-trained components are used. The speech encoder is pre-trained on the ASR task and the decoder through a MT task. In row ``II" and ``III", ST models are initialized with pre-trained encoder and decoder respectively. In both cases, more than 2 BLEU score improvements are observed. Combining both pre-trained models, another 0.2 BLEU score gain is achieved in row ``IV" compared with results in row ``II". Row ``V" is the result from the jointly trained model, which outperforms the ST model with both encoder and decoder initialized from pre-trained models by another 1.0 BLEU. 
In row ``VI", the pre-trained model is finetuned using joint training, it is slightly better than the model trained from scratch in row ``V". It shows that the jointly trained model could achieve good results even without pre-trained components.
\begin{table}[]
    \centering
    \small
    \begin{tabular}{c|c|c|c|c}
    \toprule
        model & Encoder & Decoder & Joint Training & BLEU  \\
        \hline
         I & $\times$ & $\times$ & $\times$ &  20.3 \\
         \hline
         II & $\checkmark$ & $\times$ & $\times$ & 22.7 \\
         \hline
         III & $\times$ & $\checkmark$ & $\times$ & 22.3\\
         \hline
         IV & $\checkmark$ & $\checkmark$& $\times$ & 22.9 \\
         \hline
         V & $\times$ & $\times$ & $\checkmark$ & 23.9\\
         \hline
         VI & $\checkmark$ & $\checkmark$ & $\checkmark$ & 24.0\\
          \bottomrule
    \end{tabular}
    \caption{Comparison of models initialized with pre-trained model on the MuST-C EN-DE task.}
    \label{tab:pre-train}
\end{table}

\section{Conclusion}
In this study, we propose a general multi-task learning framework to leverage text data for ASR and ST tasks. The ASR task is co-trained with a denoising autoencoder task using monolingual text, while a MT task is jointly trained with the ST task with parallel data.  Text input is represented as phoneme sequences to reduce the difference between speech input and text input. We examined different factors that impact the performance of the jointly trained system. Our experimental results show substantial WER reduction is achieved on the \textsc{Librispeech} dataset and large BLEU score gain is obtained in the \textsc{MuST-C} datasets. 
%


\bibliographystyle{IEEEbib}
\bibliography{Joint_training}
\end{document}